\documentclass[journal]{IEEEtran}

\usepackage{times}
\usepackage{epsfig}
\usepackage{graphicx}
\usepackage{amsmath}
\usepackage{amssymb}
\usepackage{longtable}
\usepackage{amsfonts}
\usepackage{algorithmic}
\usepackage{algorithm}
\usepackage{color}
\usepackage{url}
\usepackage{tabularx}

\newcommand{\minisection}[1]{\vspace{0.04in} \noindent {\bf #1}\ \ }
\begin{document}

\title{One-view occlusion detection for stereo matching with a fully connected CRF model}


\author{Mikhail~G.~Mozerov,~\IEEEmembership{~Member,~IEEE}
     and Joost~van~de~Weijer,
\IEEEcompsocitemizethanks{\IEEEcompsocthanksitem M. Mozerov and J.van de Weijer are with the Computer Vision Center
of Department Informatics, Universitat Autonoma de Barcelona, Barcelona,
Spain, 08193.\protect\\
E-mail: mozerov@cvc.uab.es}
\thanks{}}


\markboth{ ACCEPTED FOR PUBLICATION ON IEEE TRANSACTIONS ON IMAGE PROCESSING}%
{Shell \MakeLowercase{\textit{et al.}}: Improved Recursive Geodesic Distance Computation for Edge Preserving Filtering}

\IEEEcompsoctitleabstractindextext{%
\begin{abstract}
In this paper, we extend the standard belief propagation (BP) sequential technique proposed in the tree-reweighted sequential   method~\cite{Kolmogorov06} to the fully connected CRF models with the geodesic distance affinity. The proposed method has been applied to the stereo matching problem.   
Also a new approach to the BP marginal solution is proposed that we call one-view occlusion detection (OVOD). In contrast to the standard winner takes all (WTA) estimation, the proposed OVOD solution allows to find occluded regions in the disparity map and simultaneously improve the matching result. As a result we can perform only one energy minimization process and avoid the cost calculation for the second view and the left-right check procedure. We show that the OVOD approach considerably improves results for cost augmentation and  energy minimization techniques in comparison with the standard one-view affinity space implementation. We apply our method to the Middlebury data set and reach state-of-the-art  especially for median, average and mean squared error metrics.      
\end{abstract}
}
\maketitle
\IEEEdisplaynotcompsoctitleabstractindextext
\IEEEpeerreviewmaketitle
\section{Introduction}
Stereo matching is considered to be one of the fundamental problems of computer vision ~\cite{Brown03,Scharstein02}. It is important for a large variety of computer vision applications, such a intermediate view generation, 3D scene reconstruction, autonomous driving systems and robotics. 

Methods that solve the stereo matching problem can be divided into two categories: cost filtering methods  and energy minimization methods. The earliest cost filter approaches use fixed or adaptive windows as the matching neighborhood.  Later they have been extended into algorithms which operate in the bilateral color-image space~\cite{li20173d,Min11,BFS13,Sseg08,Yang12,Yoon06}. 
An efficient alternative to bilateral cost filtering is proposed by  Rhemann et al.~\cite{rhemann2011fast}, which outperforms the earliest bilateral filter in speed and quality. This  algorithm  is  based on the guided filter approach~\cite{he2013guided}. 

A common way to solve the stereo matching problem with energy minimization is by performing inference in Bayesian networks. The network should converge to the maximum a posteriori (MAP) probability or to the energy minimum of the conditional random field (CRF) functional defined over the image graph, where pixels or image patches (superpixels) are vertices and pairwise relationship between pixels are usually encoded with graph edges~\cite{kolmogorov2004energy,he2004multiscale}. Early work considers pairwise potentials of the CRF energy functional in restricted neighborhoods 
(locally connected models) ~\cite{Boykov01,Kolmogorov01,Sun03,Ihler05,Szeliski06}. This constraint makes the probabilistic graphical model less effective and flexible. 



The matching cost formation is an important component of any stereo matching algorithm.
Traditionally, costs can be divided in two groups. The first group is based on per-pixel matching dissimilarity measures~\cite{Scharstein02}. The second group is based on non-parametric transforms such as rank and census~\cite{Zabih94} or normalized cross correlation~\cite{Lewis95}. Recent progress in the field of convolutional neural networks (CNNs) has yielded more robust matching costs for stereo~\cite{zbontar2016stereo}. Consequently, the traditional cost computation has been replaced by the CNN based cost in most recent works. Also in this work cost computation is based on CNNs~\cite{taniai2016continuous}.

\subsection{Contributions}

In this paper we propose two main contributions: a) we derive the energy minimization solution for the fully connected CRF model in the framework of BP~\cite{Szeliski06} instead of mean field (MF), as used in~\cite{Krahenbuhl11}. We adapt the sequential technique proposed in~\cite{Kolmogorov06} to the fully connected CRF model, and show that the sequential algorithm for the fully connected model accelerates convergence when compared to  the non-sequential version of the algorithm; b) we propose the one-view occlusion detection (OVOD) approach that provides a new marginal solution for BP. In contrast to the standard WTA estimation the proposed OVOD solution allows to find occluded regions in the disparity map and simultaneously improves the matching result in these regions. As a consequence, we are only required to perform one energy minimization process, we can omit the cost calculation step for the second view and avoid the left-right check procedure. To demonstrate the general ability of our algorithm we apply our method to the Middlebury data set and reach state-of-the-art level for median, average and mean squared error metrics.

Note that the method proposed in~\cite{Kolmogorov06} considers only a locally-connected CRF model. In~\cite{mozerov2017improved} we proposed a recursive filter that is based on the geodesic affinity. However, this version does not apply to the energy minimization problem. Here we use that approach as a tool that allows us to extend the locally-connected CRF technique in~\cite{Kolmogorov06} to the fully-connected CRF model. Furthermore, the proposed extension of~\cite{Kolmogorov06} is theoretically proven and can be used not only in stereo but also in any other problem, which is based on energy minimization with a fully-connected CRF model. 

\subsection{Related work}
Generally cost aggregation and global optimization are considered as different techniques. Nevertheless, in two state-of-the-art works authors have proposed methods to combine local preprocessing within a global optimization framework. They combine segmentation based cost aggregation with BP energy minimization~\cite{Klaus06,Vogel11}. Firstly, an initial image is divided into relatively small segments also called superpixels. Then an aggregated cost is used for the unary potential in the energy minimization problem. There are two problems that one faces implementing this approach: firstly, segmentation is not a trivial problem especially in context of stereo matching; secondly, the assumption that pixels inside the same segment belong to similar disparity values can possibly produce estimation errors, which are impossible to correct at the stage of disparity map post-processing.

An extension  of locally connected CRF models to fully-connected CRF models 
was proposed in~\cite{mozerov2015accurate}. However in that method only one iteration is used and this iteration is equivalent to the bilateral cost filtering. An iterative solution of fully  connected CRF models was proposed in~\cite{barron2015fast}, but there the energy optimization was done by a Quasi-Newton (L-BFGS) method. 

In the segmentation paper of Krahenbuhl and Koltun~\cite{Krahenbuhl11} fully connected CRF models have been proposed. They solve the minimization problem by MF approximations. However, comparison with belief propagation  methods show that BP typically performs much better than MF~\cite{weiss2001comparing}. Thus, we propose to use the BP approach developed in the paper by Kolmogorov~\cite{Kolmogorov06}, where a sequential calculation scheme for BP convergence was developed. This method significantly reduces the number of iterations needed to reach the same level of energy compared to its non-sequential analogue proposed in~\cite{wainwright2005map}. 

The algorithms~\cite{Krahenbuhl11,barron2015fast} use the color-space bilateral distance to  compute the pairwise potential in the CRF energy functional. In the presented work the same binary potential is constructed with the geodesic distance  affinity~\cite{zhang2009cross,criminisi2010geodesic}. 
The motivation of this choice is twofold. Firstly,  fast realization of the bilateral filter is necessary to solve the CRF energy minimization problem for the fully connected model in a reasonable computational time. However, the fast bilateral filter algorithm~\cite{N21BF10}  that is used in~\cite{Krahenbuhl11,barron2015fast} is in general slower than the recursive filtering based on the geodesic distance~\cite{Gastal2011,Yang2012}, and for several kernel intrinsic parameters is considerably slower. Secondly and more important, all known fast algorithms of the bilateral filter~\cite{N14BF06,N21BF10,gastal2012adaptive,mozerov2015global} do not allow sequential update of the belief propagation messages on the image graph tree. The last observation is also relevant to the algorithms based on the guided filter approach~\cite{he2013guided}. Formally, affinity weights  in the guided filter  are calculated recursively, however such a recursion could not be adapted to the sequential process using the standard scheme in~\cite{he2013guided}.    

To realize the fast geodesic distance based filter, which is a part of our CRF energy minimization algorithm we follow the idea in~\cite{criminisi2010geodesic} and especially in~\cite{mozerov2017improved}, where recursive summation trees are proposed. Note that the algorithm in~\cite{mozerov2017improved} was developed for denoising filters. We extend this technique and adapt the recursive trees to the sequential calculation scheme for BP message passing.

In~\cite{Yang12} a non-local cost aggregation method for stereo matching was proposed, and the proposed algorithm uses the geodesic affinity for the cost filtering approach. In contrast, we use the geodesic distance not for the cost filtering, but for a message update procedure with a fully-connected CRF  model.

Now we will briefly discuss some of the most recent advances in the field of accurate stereo matching. Two recently proposed methods, the Local Expansion Moves~\cite{taniai2016continuous} and PatchMatch-based superpixel cut~\cite{li2016pmsc}, solve the energy minimization labeling problem with a locally connected graph cut technique. The methods construct the energy smoothness term that fits the solution to oriented planes, while traditional methods consider only front-parallel planes. Such a smoothness term results in increased accuracy, but increases the dimensionality of the label space from 1D to 3D, consequently increasing the general computational complexity of the methods. We discuss this further in the experimental section of our paper.  

In the paper~\cite{li20173d}, the authors propose a cost-aggregation method that embeds a minimum spanning tree based support region filtering in the general procedure. Furthermore, it combines this approach with a PatchMatch 3D label search, thus performing the search process with an adaptive patch size. An efﬁcient stereo matching algorithm, which applies adaptive smoothness constraints using texture and edge information, is proposed in~\cite{kim2016adaptive}. The approach includes two base procedures: the first one segments the disparity map domain into three classes with different smoothness terms, and the second procedure obtains disparity values by minimizing a local cost volume  energy. Finally, a 3D mesh technique is proposed in~\cite{zhang2015meshstereo}. This novel global stereo model was initially designed for view interpolation. The method estimates a disparity map and a 3D triangular mesh in such a way that both problems are solved simultaneously.


\section{Energy minimization with fully connected CRF model}\label{sec:fully}
In this section we consider energy minimization with fully connected CRF models. We extend the locally connected model proposed in~\cite{Kolmogorov06} to the fully connected one. Also we join the two step energy minimization approach~\cite{mozerov2015accurate}  (which consists of the bilateral cost filtering and the locally connected energy minimization algorithm~\cite{Kolmogorov06}) into one iterative energy minimization process of a fully connected model with the geodesic distance affinity. 

The principal difference between fully connected models and locally connected models is the number of pixels that are included in the considered pixel neighborhood.  In  fully connected models all image pixels are potentially connected to each other. However the strength of each connection depends on a chosen affinity metric between pixels. This connectivity strength is formalized by the reciprocal image affinity weights. Consequently, these weights become a part of binary potentials in the smoothness term of the energy functional. First, let us consider the energy function of the following form:
\begin{equation}
{\bf{{\rm E}}}\left( s \right) = \sum\limits_{p \in { {\cal V}}} {{c_p}\left( {{s_p}} \right)}  + \sum\limits_{\left( {p,q} \right) \in { {\cal E}}} {{B_{p,q}}\left( {{s_p},{s_q}} \right)},
\label{eq:BaseEM}
\end{equation}
where (and further in this paper) $p,q,k,l \in {\cal V}$ corresponds to pixels or vertices and set   $\varepsilon  = \left( {p,q} \right) \in {\cal E}$ to edges of an image graph $ {\cal G }= \left\{ { {\cal V}, {\cal E}} \right\}$. Where ${s_p}$ denotes the label of pixel $p$ 
that belongs to some discrete set of disparities  $s_p \in S$;   ${c_p}\left(  \cdot  \right)$ 
defines a unary potential which corresponds to the conventional penalty cost or the logarithm of the likelihood probability. 
${B_{p,q}}\left( { \cdot , \cdot } \right)$   is a binary potential which defines edge interaction between pixels $\left( {p,q} \right)$.
In the case of the fully connected CRF model each combination of pixels   
$\left( {p,q} \right)$ of the image graph   ${\cal G}$ belongs to the edge set  ${\cal E}$. And the binary potential for the model can be expressed as:
\begin{equation}
{B_{p,q}}\left( {{s_p},{s_q}} \right) = {b_q}{w_{p,q}}\varphi \left( {{s_p},{s_q}} \right),
\label{eq:FullM}
\end{equation}
where  $\varphi \left( { \cdot , \cdot } \right)$ is a pairwise penalty that enforces smoothness 
of the estimated label function, and ${w_{p,q}}$ is a pairwise weight that defines some reciprocal influence between every node. 
The variable ${b_q}$ is a  normalization factor set to
\begin{equation}
{b_q}={\left( {\sum\nolimits_{p \in \mathcal V} {{w_{p,q}}} } \right)^{ - 1}}
\end{equation}

The problem of energy minimization in Eq.~(\ref{eq:BaseEM}) can be solved in the framework of BP. The aim is to find the marginal function  ${\bar c_q}\left(  \cdot  \right)$ and then estimate labels in each pixel $q$ by the simple  WTA formula, which finds the maximum likelihood:
\begin{equation}
s_q = \mathop {\arg \min }\limits_{s \in S} {\bar c_q}\left( s \right)
\label{eq:wta}
\end{equation}

The core operation of BP is passing a message  ${m_{p \to q}}$ from node $p$ to node $q$  for the directed edge   $\left( {p \to q} \right) $,  where $\left( {p , q} \right) \in { {\cal E}}$. The marginal function  is then expressed as follows:
\begin{equation}
\bar c_q^t\left( s \right) = {c_q}\left( s \right) + \sum\limits_{p, q \in { {\cal E}}} {m_{p \to q}^{t - 1}\left( s \right)},
\label{eq:mrgnl}
\end{equation}
or the following
\begin{equation}
\bar c_q^t\left( s \right) = {c_q}\left( s \right) + \sum\limits_{p|p, q \in { {\cal E}}} {m_{p \to q}^{t - 1}\left( s \right)}
\label{eq:mrgnl2_}
\end{equation}
Here $t$ is the number of the iteration, and it is supposed that the related iterative message passing procedure is given~\cite{Szeliski06}:
\begin{equation}
m_{p \to q}^t\left( i \right) = \mathop {\min }\limits_{j \in S} \left( {{\beta _{p,q}}\bar c_p^t\left( j \right) - m_{p \to q}^{t - 1}\left( j \right) + {B_{p,q}}\left( {i,j} \right)} \right),
\label{eq:mssg}
\end{equation}
where indexes $ {i,j}  \in S$ belong to the discrete disparity domain $S$, 
 and ${\beta _{p,q}}$ is a normalization 
factor that is set to ${{\beta _{p,q}} = {b_q}{w_{p,q}}}$. The term $m_{p \to q}^{t - 1}\left( j \right)$ 
can be neglected for the fully connected model, because in general this value vanishes  relative to the sum of all other messages incoming to pixel $q$. 
Thus  we can rewrite Eq.~(\ref{eq:mssg}) as follows:
\begin{equation}
\begin{array}{l}
 m_{p \to q}^t\left( i \right) = {b_q}{w_{p,q}}m_p^t\left( i \right), \\ 
 m_p^t\left( i \right) = \mathop {\min }\limits_{j \in S} \left( {\bar c_p^t\left( j \right) + {\varphi _{p,q}}\left( {i,j} \right)} \right) \\ 
 \end{array}
\label{eq:mssg2}
\end{equation}
The marginal of Eq.~(\ref{eq:mrgnl})  can also be rewritten:
\begin{equation}
\bar c_q^t\left( s \right) = {c_q}\left( s \right) + \frac{b_q}{{1- b_q  }}\sum\limits_{p \in {\cal V}} {{w_{p,q}}m_p^{t - 1}\left( s \right)}  - m_q^{t - 1}\left( s \right)
\label{eq:mrgnl2}
\end{equation}

One can see that the sum term  ${\sum\nolimits_{p \in {\cal V}} {{w_{p,q}}m_p^{t - 1}} }$ in Eq.~(\ref{eq:mrgnl2}) coincides with the discrete image filtering definition for a large class of filter kernels ${{w_{p,q}}}$, which include traditional Gaussian, bilateral  or geodesic distance based kernels. In this paper we propose to use the fast geodesic distance based filter described in the next Section, because its recursive graph-tree based calculation allows sequential updating of messages using Eq.~(\ref{eq:mssg2}). Thus the solution of the energy minimization problem with the CRF fully connected model can be obtained by an iterative process using Eq.~(\ref{eq:mrgnl2}), Eq.~(\ref{eq:mssg2}), and Eq.~(\ref{eq:wta}). 

\section{Geodesic distance based recursive calculation tree}

\begin{figure}[t]
\centering
\includegraphics[width=0.4\textwidth]{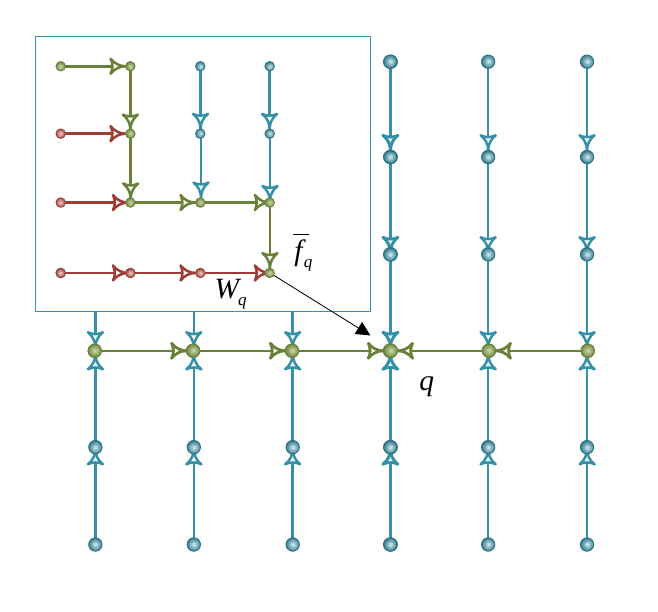}
\caption {One of four recursive calculation trees~\cite{mozerov2017improved}.}
\label{fg:tree}
\end{figure}
As  pointed out in Section II we propose to use the fast geodesic distance, because its recursive calculation allows sequential updating of messages. Consequently this leads to faster and more accurate convergence of the energy minimization process in contrast to the bilateral or the guided filter. In addition, the topology of the standard bilateral filter with Gaussian kernel does not fit well to the stereo matching problem. This problem arises because disparity   values of two different but near pixels can considerably differ while belonging to the same color of a cover image. In this case, the bilateral filter blurs both output values.   This ability of the bilateral filter to communicate over color edges could be useful for example for color based segmentation problem, but can produce estimation artifacts for stereo. Consequently, we found that edge preserving filters based on the geodesic distance measure are more suitable to the stereo matching problem.

The geodesic distance based filter is usually chosen in the following form, which makes the filter recursive
\begin{equation}
\begin{array}{*{20}{c}}
   {{{\tilde f}_q} = \frac{1}{{{W_q}}}\sum\limits_{p \in {\cal V}} {{e^{ - a{d_{p,q}}}}} {f_p}}  \\
   {{W_q} = \sum\limits_{p \in {\cal V}} {{e^{ - a{d_{p,q}}}}} },  \\
\end{array}
    \label{eq:filter_base}
\end{equation}
where$f_p$ and ${{{\tilde f}_q}}$ are the input and the output of the filter respectively,  the weight  $ e^{ - a{d_{p,q}}} $  defines a geodesic distance based affinity between any two image pixels. 

The variable  ${d_{p,q}}$ in Eq.~(\ref{eq:filter_base}) is the geodesic distance between image pixels $\left( {p,q} \right)$ which for an image ${I_p}$ can be defined on the discrete grid graph as 
\begin{equation}
\begin{array}{l}
 {d_{q,p}} = \mathop {\min }\limits_{{P_{p,q}}}\; \sum\limits_{\varepsilon  \in {P_{p,q}}} {{u_\varepsilon }} , \\ 
 {u_{\varepsilon  = \left( {q,p} \right)}} = \left\| {{I_q} - {I_p}} \right\| + \delta , \\ 
 \end{array}
\label{eq:GD_def}
\end{equation}
where ${P_{q,p}}$ is a path between two graph vertices $\left( {p,q} \right)$ and $\delta $ is the space term. Remember that in graph theory, a path in a graph is a  sequence of edges which connect a sequence of vertices.

Note that the filter intrinsic parameters $a$ and $\delta $  in Eq.~(\ref{eq:GD_def}) 
approximately correspond to the parameters of the classic bilateral filter with the Gaussian kernel as follows
\begin{equation}
a = \frac{2}{{\sigma _r^2}},{\rm{ }}\delta  = \frac{{\sigma _r^2}}{{\sigma _s^2}},
    \label{eq:K_par}
\end{equation}
where $\sigma _r^2$ and $ \sigma _s^2$ are the range and the space variance respectively.
Also from Eq.~(\ref{eq:GD_def}) one can derive the following  useful relation
\begin{equation}
{w_{p,q}} = {e^{\mathop {\min }\limits_{{P_{p,q}}} \sum\limits_{\varepsilon  \in {P_{p,q}}} {{u_\varepsilon }} }} = \mathop {\max }\limits_{{P_{q,p}}} \prod\limits_{\varepsilon  \in {P_{p,q}}} {{e^{ - a{u_\varepsilon }}}}  
\label{eq:PR_def}
\end{equation}
In~\cite{mozerov2017improved} it is shown that the filter sum in Eq.~(\ref{eq:mrgnl2}) can be calculated efficiently and this calculation scheme allows for  sequential message updates.

The full tree, which corresponds to this fast algorithm~\cite{mozerov2017improved}, is composed by four quadrant-domains (or branches of the tree). One of four branches is illustrated in Fig.~\ref{fg:tree}. Here
$W_q$ represents the sum of weights  $w_{p,q}$   for the first quadrant, when ${{\tilde f}_q}$ gives filter results also for this quadrant. 
Note that the optimal tree and all weights in Eq.~(\ref{eq:PR_def}) can be calculated  for all other procedures. 

\subsection{Message sequential update}
The difference between sequential and iterative BP message updates is defined in ~\cite{Kolmogorov06}. According to this definition one can rewrite the marginal Eq.~(\ref{eq:mrgnl2}) as
\begin{equation}
\bar c_q^t\left( s \right) = {c_q}\left( s \right) + \frac{1}{{{W_q} - 1}}\sum\limits_{p \in {\cal V}} {{w_{p,q}}m_p^{t}\left( s \right)}  - m_q^{t}\left( s \right)
\label{eq:mrgnl3}
\end{equation}
Taking into account Eq.~(\ref{eq:mrgnl}) one can say that a message coming from pixel $p$ depends on all incoming messages of the same iteration. However, because the sequential process assumes that the message calculation comes recursively on  the tree branch (quadrant-domain) from leaves (all previous pixels of the branch) to root $p$, only one branch update matters and the full iteration is composed of all four branch passes.

To provide more intuition on the sequential idea of BP message updating, let us consider the one dimensional case (e.g. a string in an image) of BP convergence. It is known that the energy converges after finite number of iterations and that this number is proportional or equal to the number of nodes of which the graph of this one dimensional data consists~\cite{felzenszwalb2006efficient}. On the other hand, a dynamic programming (DP) approach can achieve the solution at once, which is equivalent to one iteration. Note that a recursive transition function from vertex to vertex of DP in this case is exactly equivalent to the BP message coming from the same vertices.  Thus the proposed sequential approach applies this dynamic programming principle to the belief propagation technique. 

\section{One-view occlusion detection concept}
\begin{figure}[t]
\centering
\includegraphics[width=0.45\textwidth]{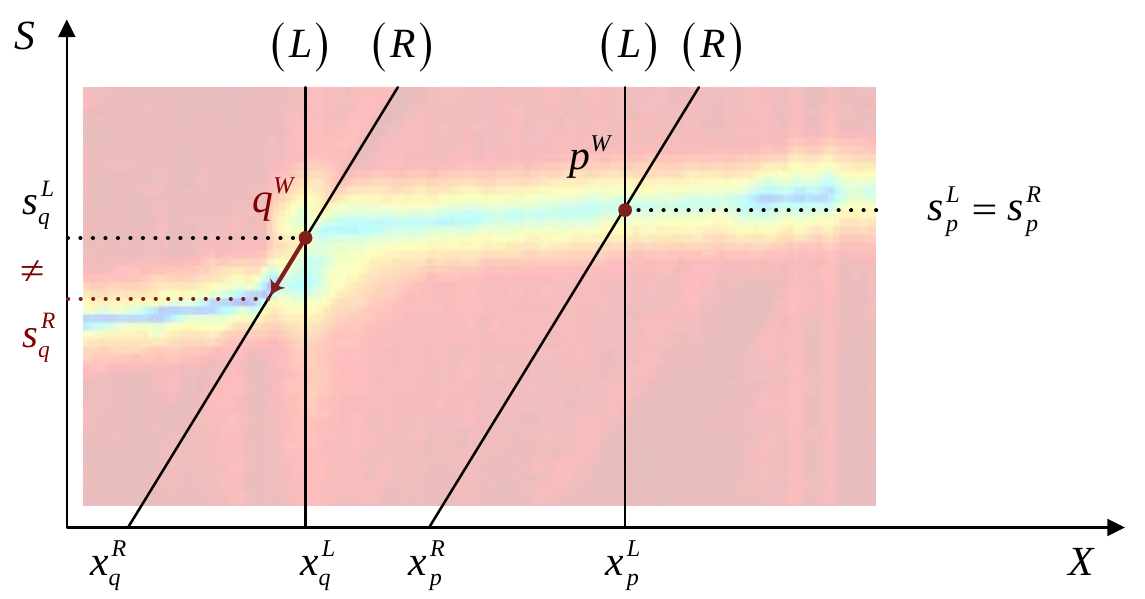}
\caption {Two cases of left-right marginals intersection:  for the non-occluded pixel $p$, where $s_p^L = s_p^R$; and for the occluded pixel $q$, where $s_q^L \ne s_q^R$.}
\label{fg:ovod}
\end{figure}
Before we explain the idea of one-view occlusion detection (OVOD), we briefly describe the base approach that detects occlusion in stereo and that is used in almost all state-of-the-art algorithms. It is the so called left-right  consistency check~\cite{egnal2002detecting}. This technique involves two independent cost volume calculations. Our method allows to use only one cost volume and therefore only one energy minimization procedure is necessary. 

The left-right  consistency check approach supposes that firstly two independent cost volumes 
$c^L(q,s)$ and $c^R(q,s)$ have to be calculated, where pixels  ${q^L}$ and ${q^R}$  can be represented as two corresponding points with coordinates in the left ${q^L} = {\left[ {x_q^L,y_q^L} \right]^T}$ and the right ${q^R} = {\left[ {x_q^R,y_q^R} \right]^T}$ image planes respectively. Note that here ${s^L}$ and ${s^R}$ are disparity coordinates of the left and the right view cost volumes respectively. Then both cost volumes are usually processed by energy minimization or cost filtering algorithms to obtain marginal $\bar c^L(q,s)$  and $\bar c^R(q,s)$ . Consequently, two independent solutions $s_q^L$ and $s_q^R$ can be obtained by using Eq.~(\ref{eq:wta}).

Finally, consistency check formula is used to detect occluded pixels:
\begin{equation}
{o_{{q^L}}} = \left\{ \begin{array}{l}
 \begin{array}{*{20}{c}}
   {occ} & {{\text{if }}{s^L}\left( {x_q^L,{y_q^L}} \right) \ne {s^R}\left( {x_q^L - s_q^L,{y_q^L}} \right)}  \\
\end{array} \\ 
 \begin{array}{*{20}{c}}
   {nocc} & {{\text{elsewhere}}}  \\
\end{array} \\ 
 \end{array} \right.
\label{eq:ov1}
\end{equation}
where ${o_{{q^L}}}$ is the consistency check  output occlusion map. If a value ${o_{{q^L}}} = occ$ it means that an occluded pixel is detected, and the previously estimated value $s_q^L$ is uncertain and should be replaced by any gap-filling post-processing procedure.

The idea of OVOD is to avoid the double-cost volume procedure and to omit the gap-filling procedure. It is based on the observation that any visible 3D world point ${q^W}$ can be mapped into both cost volume coordinate systems by ${q^W} \Rightarrow \left[ {x_q^L,y_q^L,s} \right]$ and ${q^W} \Rightarrow \left[ {x_q^R,y_q^R,s} \right]$ with clear relationship between both mapping: $x_q^L - x_q^R = s$ and $y_q^L = y_q^R$. A cost volume value  ${e^{ - c\left( {{q^W}} \right)}}$ defines a probability of the world point ${q^W}$ to belong to a point on the surface of the object. The point ${q^W}$ can be mapped into both left and right cost volumes with the same cost value $c\left( {{q^W}} \right)$.  Thus, one can assume that it is possible to use the left-view cost volume as the right-view cost volume after a simple shearing  transformation that changes only the X-coordinate:
\begin{equation}
x_q^R = x_q^L - s 
\label{eq:ov0}
\end{equation}
Then the left-view marginal volume $\bar c^L(q,s)$   also can be considered as the right-view marginal volume   $\bar c^R(q,s)$  but with the transformed X-coordinate given by Eq.~(\ref{eq:ov0}). In this case the right-view solution $s_q^R$ is obtained by modified Eq.~(\ref{eq:wta}):
\begin{equation}
s_q^R = \mathop {\arg \min }\limits_{s \in S} \bar c\left( {x_q^L - s,{y_q},s} \right)
\label{eq:ov2}
\end{equation}

As is  illustrated  in Fig.~\ref{fg:ovod}  a marginal solution $s_q^L$ for a fixed pixel ${q^L}$ is found at the minimum in the vertical direction (L) of the unified marginal volume. Another solution  is found at the minimum in the slanted direction (R) defined in Eq.~(\ref{eq:ov2}) and illustrated in Fig.~\ref{fg:ovod}. To give an intuition: one can compare the (L) and the (R) directions with light rays coming from a cost volume mapped world point ${p^W}$ (see Fig.~\ref{fg:ovod}) to the left and to the right lenses of a stereo system respectively.  If the value of ${o_{{q^L}}} = nocc$ it means that both WTA minimization  directions (L)  in Eq.~(\ref{eq:wta}) and (R)  in Eq.~(\ref{eq:ov2}) should intersect in the solution point $s_q^L = s_q^R$ . Therefore, when $s_q^L = s_q^R$  this means that an occluded pixel is detected, or the world point ${q^W}$ does not belong to the reflected surface of any object. In stereo matching the occluded pixel  should be treated. We use the following simple occlusion handling to avoid a gap-filling procedure:
\begin{equation}
{s_q} = \min \left( {s_q^L,s_q^R} \right)
\label{eq:ov3}
\end{equation}

The idea behind this choice is simple and illustrated in Fig.~\ref{fg:ovod}: if pixel $x_q^L$ is occluded then the neighboring pixel to the right in Fig.~\ref{fg:ovod} is also occluded or its disparity value is larger than $s_q^L$. Note that we cannot use larger values to fill the gap due to the occlusion definition: disparity of an occluded pixel should always be less than the disparity of pixels that occlude. Thus, to replace the uncertain value $s_q^L$ we have to choose the disparity value from the left neighborhood pixels and Eq.~(\ref{eq:ov3}) satisfies the proposed assumption (see Fig.~\ref{fg:ovod}).   However, it is possible to use other algorithms to fill the occluded regions, because in our algorithm the occlusion map ${o_{{q^L}}}$ is calculated as a by-product of the main procedure.    
\section{Stereo matching algorithm}\label{sec:Stereo}
In this section we describe the application of our theory for the specific stereo matching problem. 

\subsection{Energy potentials}
Firstly, one has to define the main functional in Eq.~(\ref{eq:BaseEM}) for the stereo matching specific problem. The unary potential Eq.~(\ref{eq:BaseEM}) or the cost variable ${c_p}\left( {{s_p}} \right)$ in our experiments is directly incorporated from  the state-of-the-art CNN-based matching cost function by~\cite{taniai2016continuous}. This cost is computed by the approach proposed in~\cite{zbontar2016stereo}. The CNN-cost  calculation function $ c_p^{CNN}\left( {p,p - s} \right)$ computes a matching cost between left and right image patches of size $11\times11$ centered at $p$ and $p-s$, respectively, using the MC-CNN neural network~\cite{zbontar2016stereo}. The cost values relevant to the Middlebury data set~\cite{scharstein2014high} (half-size) are directly taken from the web site of the project~\cite{taniai2016continuous} (see relevant reference in the Middlebury evaluation table). Note that in our experiments we  use only the left-to-right cost. We use truncated version of this cost $c_p^{CNN}$:
\begin{equation}
{c_p} = \min \left( {1.5c_p^{CNN},\sqrt {c_p^{CNN}} } \right)
\label{eq:costCNN}
\end{equation}
The binary potential ${B_{p,q}}\left( { \cdot , \cdot } \right)$  which is redefined via $\varphi \left( {{s_p},{s_q}} \right)$ in Eq.~(\ref{eq:FullM}) is a truncated linear function:
\begin{equation}
\varphi \left( {{s_p},{s_q}} \right) = \lambda \min \left( {\left| {{s_p} - {s_q}} \right|,{\Delta _{\max }}} \right),
\label{eq:Bpotential}
\end{equation}
where $\lambda$ in our experiments is set to 0.12 and the base value of $\Delta _{\max }$ is 8.
For an additional experiment to compare our approach with similar method~\cite{mozerov2015accurate} we use the cost calculating algorithm taken from that paper. 

 \subsection{Image-cost downscaling}
 To make our experiments tractable on a CPU computer we perform a downscale procedure for the cost and stereo images, which are used to form the CRF model affinity space. In general we use different scale factor for the different size of the input image. The base values of the scale factors in our experiments are: $M=4$. There are exceptions for the stereo images with the disparity range  $S>300$: $M=5$ and for images with full-size less than $1000 \times 500$: $M= 3$.  Downscaling is performed with the raised cosine kernel, which weight function is:

\begin{equation}
{w^{RC}}\left( {z,p} \right) = \left\{ \begin{array}{l}
 \begin{array}{*{20}{c}}
   0 & {{\rm{if }}\left\| {zM - p} \right\| > 1}  \\
\end{array} \\ 
 \begin{array}{*{20}{c}}
   {\frac{1}{2}\left( {1 + \cos \left( {\frac{{\pi \left\| {zM - p} \right\|}}{{M}}} \right)} \right)} & {{\rm{else}}}  \\
\end{array}, \\ 
 \end{array} \right.
\label{eq:dwsc}
\end{equation}
where $z$ is a pixel on the downscaled image grid, and $p$ is a pixel on the original grid.

For the cost volume we also perform downscaling in the label-disparity domain with the factor $M$, but in this case we take the min-value of the cost inside the scale interval and remember the real disparity value of this minimum to recover disparity values with initial cost accuracy.  

\subsection{Fully connected model parameters}
To solve the functional in Eq.~(\ref{eq:BaseEM}) and thus solve the main part of our stereo matching algorithm one needs to perform the message sequential update process defined in Eq.~(\ref{eq:mrgnl3}). To do that we need to define intrinsic metric weights $w_{p,q}$  of our fully connected model. These weights $w_{p,q}$ are calculated with Eqs.~(\ref{eq:GD_def}-\ref{eq:K_par})   
and in our experiments $\sigma _r =30$ and $ \sigma _s=8$ are the range and the space standard deviation  of the model affinity space respectively.

 \subsection{Disparity map upscaling}
 Because the output results ${s^{out}}\left( {x,y} \right)$ should be the same size as an input image it is necessary to upscale the disparity map ${s^{dwsc}}\left( {\xi ,\zeta } \right)$  of the energy minimization on the downscaled grid $\left( {\xi ,\zeta } \right)$ of the main process.  For this purpose we propose to use the geodesic distance filter with further notation ${F^{GD}}\left\{  \cdot  \right\}$   that described in Eq.~(\ref{eq:filter_base}). Firstly, let us define the space functions $s^{in}$ and $msk^{in}$ that are input for the the filter ${F^{GD}}\left\{  \cdot  \right\}$:
\begin{equation}
\begin{array}{l}
 {s^{in}}\left( {x,y} \right) = \left\{ \begin{array}{l}
 \begin{array}{*{20}{c}}
   {{s^{dwsc}}\left( {\xi ,\zeta } \right)} & \begin{array}{l}
 x = M\xi  \\ 
 y = M\zeta  \\ 
 \end{array}  \\
\end{array} \\ 
 \begin{array}{*{20}{c}}
   0 & \text {elsewhere}  \\
\end{array} \\ 
 \end{array} \right. \\ 
 ms{k^{in}}\left( {x,y} \right) = \left\{ \begin{array}{l}
 \begin{array}{*{20}{c}}
   1 & \begin{array}{l}
 x = M\xi  \\ 
 y = M\zeta  \\ 
 \end{array}  \\
\end{array} \\ 
 \begin{array}{*{20}{c}}
   0 & \text {elsewhere}  \\
\end{array} \\ 
 \end{array} \right. \\ 
 \end{array}
\label{eq:upscl2}
\end{equation}
Note that here the variables $\xi$ and $\zeta$ belong to the integer values.

Then we apply filter to the input functions $s^{in}$ and $msk^{in}$ to obtain 
the  output results ${s^{out}}\left( {x,y} \right)$:
\begin{equation}
\begin{array}{l}
 {s^{out}} = \frac{1}{{{W^{out}}}}{F^{GD}}\left\{ {{s^{in}}|{I^L},{\sigma _{ru}},{\sigma _{su}}} \right\} \\ 
 {W^{out}} = {F^{GD}}\left\{ {ms{k^{in}}|{I^L},{\sigma _{ru}},{\sigma _{su}}} \right\}, \\ 
 \end{array}
\label{eq:upscl}
\end{equation}
where $I^L$ is the left stereo image that defines the geodesic distance affinity of the filter in Eq.~(\ref{eq:filter_base}),
and ${\sigma _{ru}}$ and ${\sigma _{su}}$ are intrinsic parameters of the filter relevant 
to Eq.~(\ref{eq:K_par}). In our experiments  ${\sigma _{ru} = 0.85M}$ and ${\sigma _{su}=0.85M}$.

\subsection{Disparity map post-processing}
To obtain state-of-the-art results on stereo estimation several post-processing steps are required. We briefly summarize them here.

\minisection {Weighted median filter} Note that this filtering could be done  before upscaling.  
Weighted median filter usually is a robust extension of the bilateral filter and widely used in stereo matching post-processing, for example in~\cite{Yang09}. The main idea is to accumulate a weighted histogram ${h_q}\left( i \in S \right)$ in every pixel based on a previously estimated disparity map:

\begin{align}
{h_q}\left( i \right) = \sum\limits_{p \in {N_\rho }} {{e^{ - \frac{{{{\left| {p - q} \right|}^2}}}{{2{\rho ^2}}} - \frac{{{{\left| {{I_p} - {I_q}} \right|}^2}}}{{2\sigma _m^2}}}}\delta_K \left( {{s_p},i} \right){s_p}},
\label{eq:med}
\end{align}
where ${{s_p}}$ is a value of the disparity map obtained in the previous step, $\delta_K$ is the Kronecker delta function,  $N_\rho$ is a squared neighborhood window of size $\left(2 \rho +1\right) \times \left(2 \rho + 1\right)$, and ${\sigma _m}$ and ${ \rho}$ are two intrinsic parameters of the weighted median filter. We take the  value of  ${ \rho = 3}$  and  ${\sigma _m = 10}$ for not occluded region, and  ${ \rho = 8}$  and  ${\sigma _m = 10}$ for occluded. 
Then the new value of the desired disparity map is estimated as follows:
\begin{align}
\mathop {{s_q} = \arg {\mathop{\rm med}\nolimits} }\limits_{i \in S} \left( {{h_q}\left( i \right)} \right)
\label{eq:med2}
\end{align}

\minisection {Plane fitting filter} The filter exploits the assumption that the desired disparity function $s_q$ belongs locally to the geometric plane  with constant gradient ${\nabla {s_q}}$. The  idea of this filter is partially taken from~\cite{taniai2016continuous}, and we define it as: 
\begin{align}
{{\tilde s}_q} = \sum\limits_{p \in {N_\rho }} {{e^{ - \frac{{{{\left| {{\pi _p} - {s_q}} \right|}^2}}}{{2\sigma _\pi ^2}} - \frac{{{{\left| {\nabla {s_p} - \nabla {s_q}} \right|}^2}}}{{2\sigma _g^2}}}}\pi _p\left( {\nabla {s_p},{s_p},q} \right),} 
\label{eq:plf}
\end{align}
 where ${{\tilde s}_q}$ is the output of the filter and 
 ${\pi _p} = \pi _p\left( {\nabla {s_p},{s_p},q} \right)$ is a simple predictor function, which can predict value of $s_q$ with a known value of the gradient  $\nabla {s_p}$ and the value ${s_p}$ at the pixel $q$. The intrinsic parameter ${\sigma _\pi}$ defines the affinity between the predicted value of ${\pi _p}$ and the given value  $s_q$  and it is equal to 5; another intrinsic parameter of the filter $\sigma _g = 0.2$
enforces the weight $\exp \left( { - {{\left| {\nabla {s_p} - \nabla {s_q}} \right|}^2}/2\sigma _g^2} \right)$ to be zero if the gradients of the desired function $s$ in pixels $q$ and $p$ are considerably different, and thus belong to the different planes; ${{N_\rho }}$ is the local neighborhood window of size $15 \times 15$. 

\minisection {Disparity sub-pixel enhancement} For this post-processing 
step we use the approach realized in~\cite{Yang09}. The method interpolates the initial cost near the base disparity map with a quadratic polynomial, and then finds the local minimum of the disparity in each pixel.

\minisection {Small area outliers suppression} This post-processing 
step approach was realized in~\cite{mozerov2015accurate}. The idea is to detect outliers by  segmenting the output disparity map and then finding regions with relatively small area. Consequently, the detected regions are filled by neighborhood values.

Note that the choice of post-processing methods and full pipeline intrinsic parameters of our algorithm is motivated by the output accuracy results based on the Middlebury v3 training data set processing. Our method post-processing steps are applied in in the same order as it is described in  this Section.   
Our readers can find more technical details  inside  our C++ project for Microsoft Visual Studio 2017 that is referred in the Middlebury evaluation table.   

\section{Experiments}\label{sec:experiments}

\begin{figure*}[t]
\centering
\begin{tabular}{c}
\includegraphics[width=0.95\textwidth]{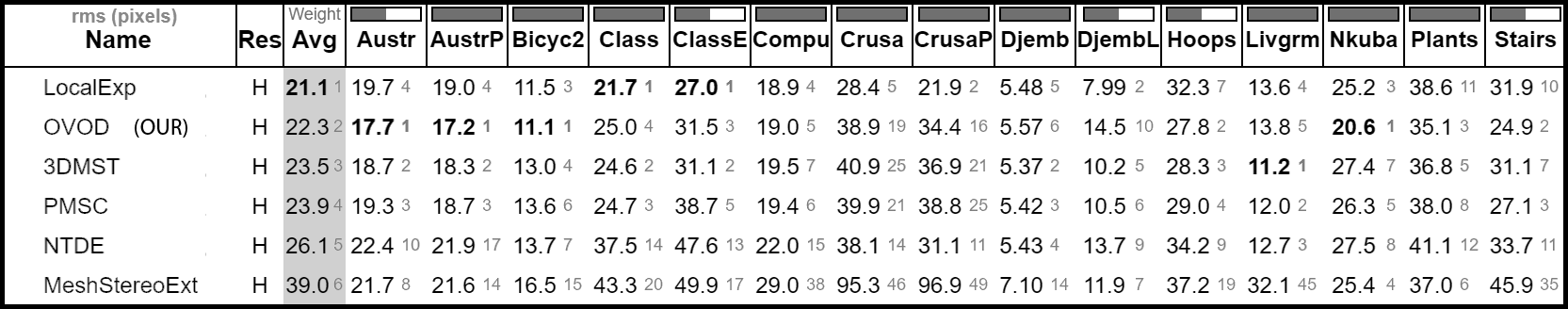} \\
(a)  \\
 \includegraphics[width=0.95\textwidth]{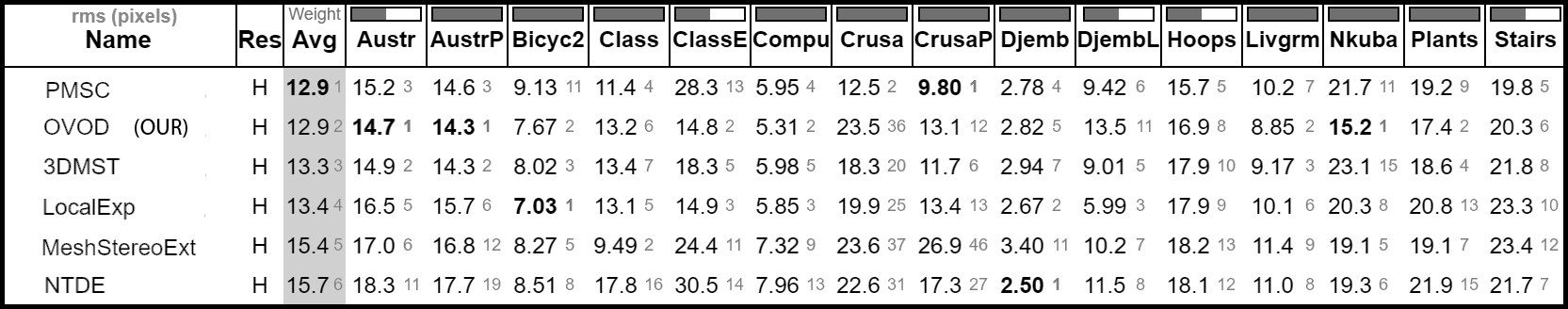} \\
(b)  \\
\end{tabular}
\caption {Middlebury benchmark V3 for the root-mean-squared error metric. Here we compare our method with the other top five methods (Snapshot from website on December 15, 2017): (a) - results for the all pixels mask; (b) - results for the non-occluded mask. Note that our method is second in both Tables and that the first position is taken by two different methods. Among the two Tables we obtain first position for 7 images, more than any of the other methods.}
\label{fg:table}
\end{figure*}
\begin{figure*}[t]
\centering
\begin{tabular}{c}
\includegraphics[width=0.95\textwidth]{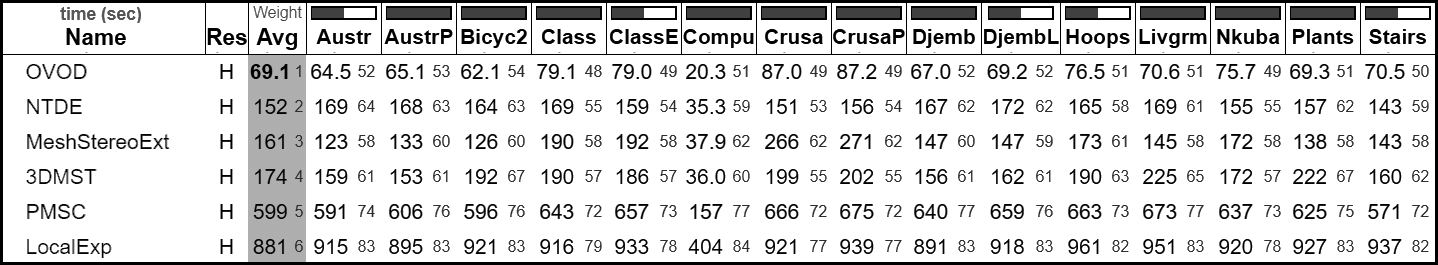} \\
\end{tabular}
\caption {Middlebury benchmark V3 for the run-time metric. Here we compare our method with the other top five methods (Snapshot from website on December 15, 2017). Note that our method is on the first position  then the others  are in inverse order in comparison with the results that are shown in  Fig.~\ref{fg:table}, where methods accuracy is considered. Here the unit  of time is CPU seconds.   }
\label{fg:table_time}
\end{figure*}
\begin{figure*}
\begin{center}
\begin{tabular}{cccccc}
Images&
\includegraphics[width=0.31\columnwidth]{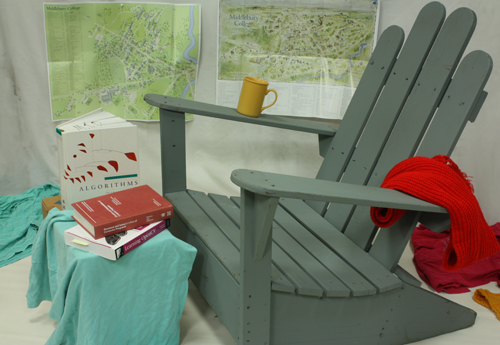} &
\includegraphics[width=0.32\columnwidth]{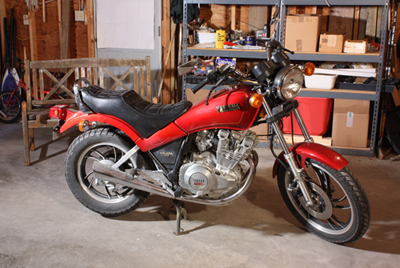} &
\includegraphics[width=0.313\columnwidth]{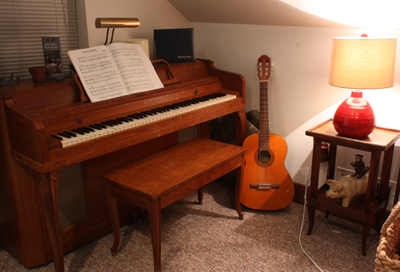} &
\includegraphics[width=0.318\columnwidth]{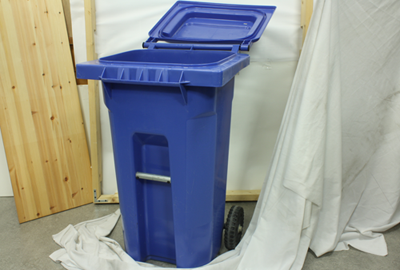} &
\includegraphics[width=0.258\columnwidth]{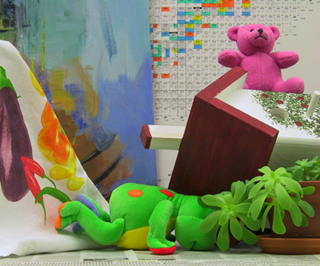} \\
GT&
\includegraphics[width=0.31\columnwidth]{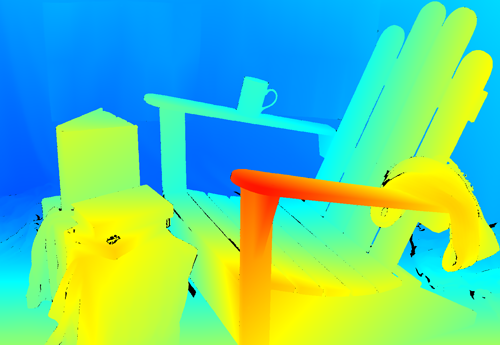} &
\includegraphics[width=0.32\columnwidth]{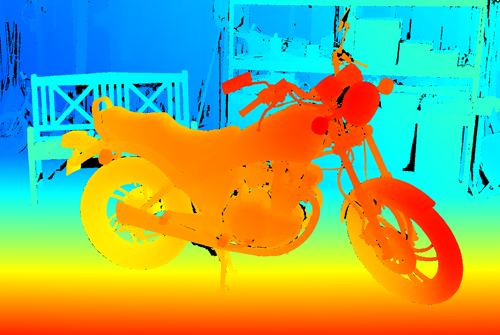} &
\includegraphics[width=0.313\columnwidth]{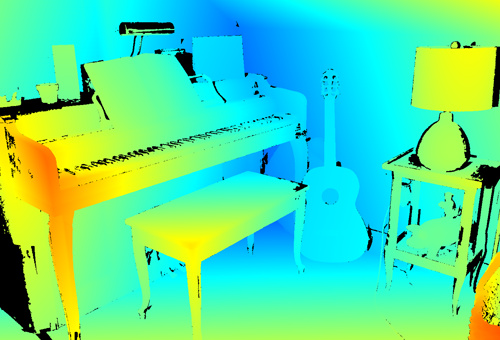} &
\includegraphics[width=0.318\columnwidth]{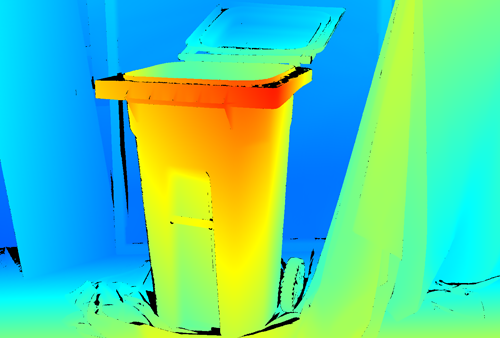} &
\includegraphics[width=0.258\columnwidth]{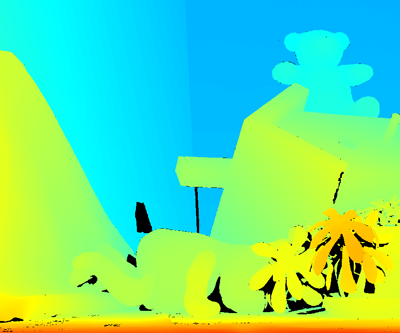} \\
OVOD~(our)&
\includegraphics[width=0.31\columnwidth]{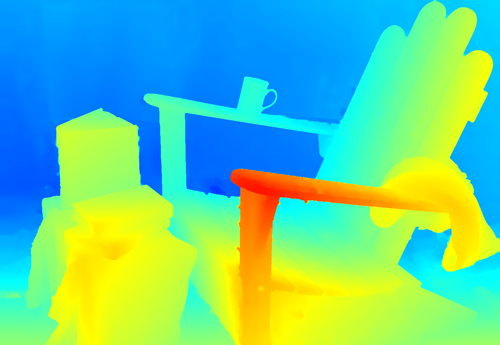} &
\includegraphics[width=0.32\columnwidth]{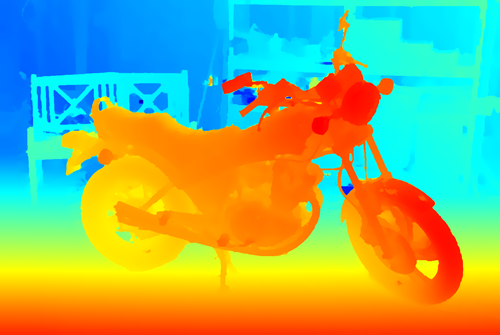} &
\includegraphics[width=0.313\columnwidth]{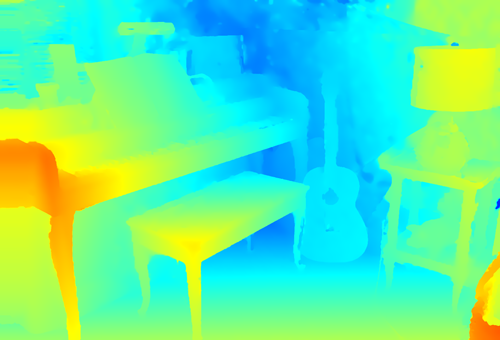} &
\includegraphics[width=0.318\columnwidth]{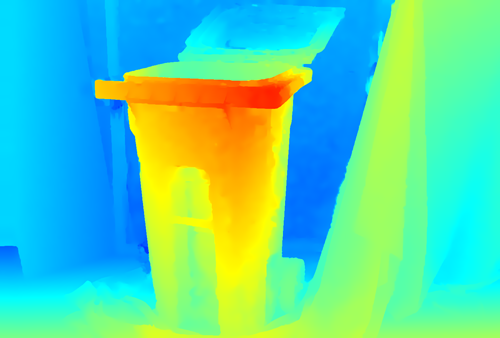} &
\includegraphics[width=0.258\columnwidth]{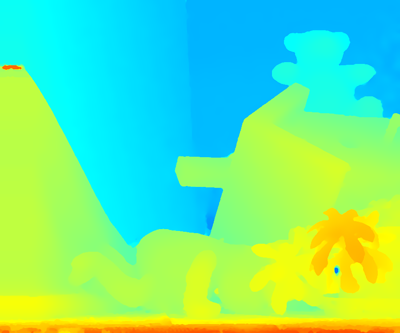} \\
LocalExp~\cite{taniai2016continuous}&
\includegraphics[width=0.31\columnwidth]{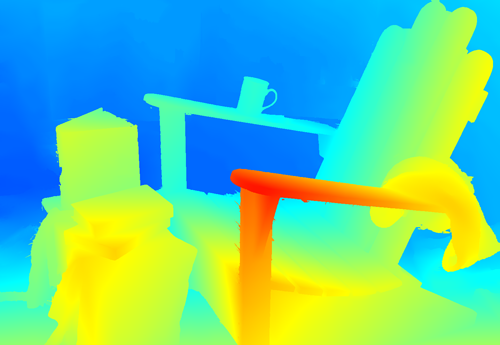} &
\includegraphics[width=0.32\columnwidth]{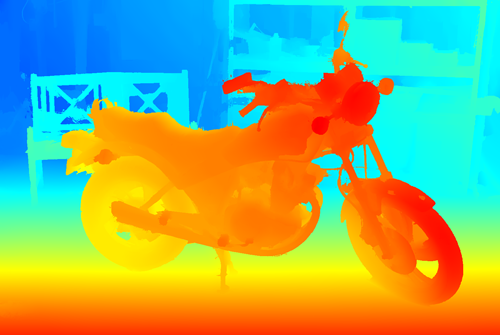} &
\includegraphics[width=0.313\columnwidth]{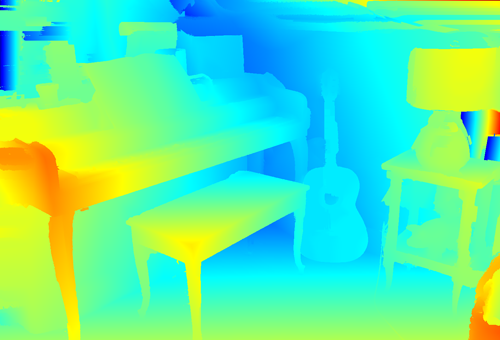} &
\includegraphics[width=0.318\columnwidth]{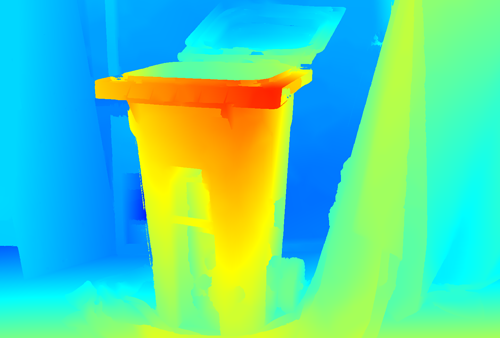} &
\includegraphics[width=0.258\columnwidth]{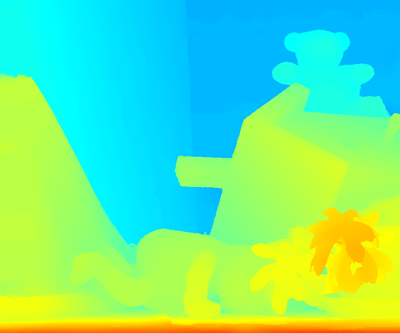} \\
3DMST~\cite{li20173d}&
\includegraphics[width=0.31\columnwidth]{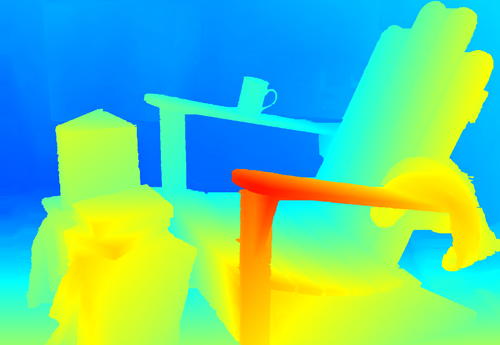} &
\includegraphics[width=0.32\columnwidth]{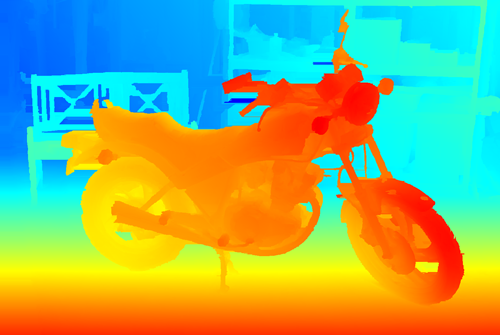} &
\includegraphics[width=0.313\columnwidth]{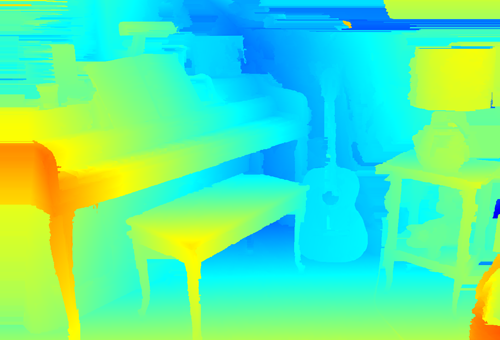} &
\includegraphics[width=0.318\columnwidth]{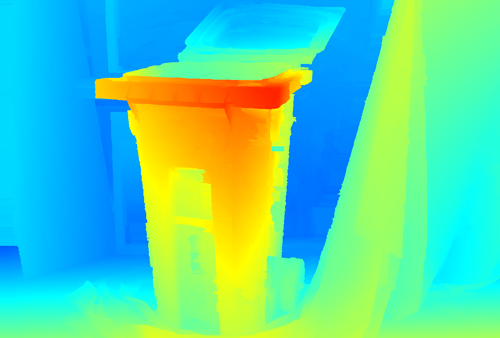} &
\includegraphics[width=0.258\columnwidth]{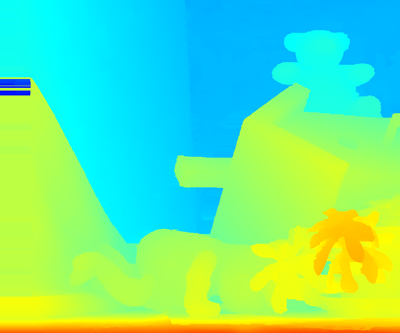} \\
 PMSC~\cite{li2016pmsc}&
\includegraphics[width=0.31\columnwidth]{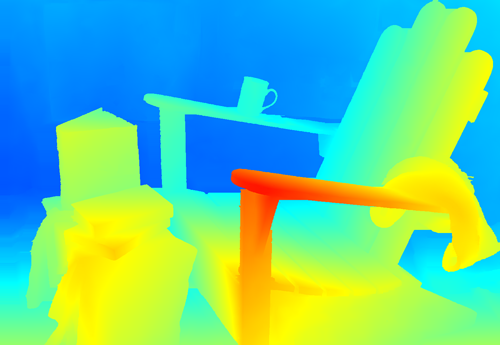} &
\includegraphics[width=0.32\columnwidth]{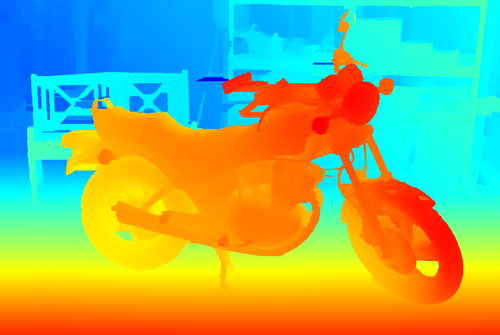} &
\includegraphics[width=0.313\columnwidth]{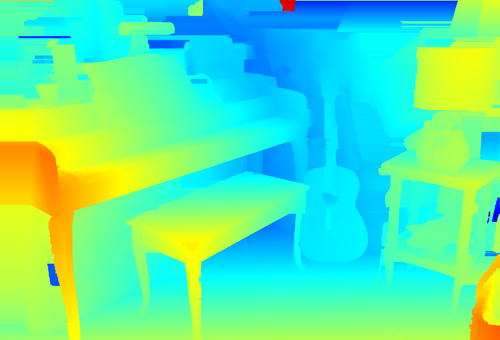} &
\includegraphics[width=0.318\columnwidth]{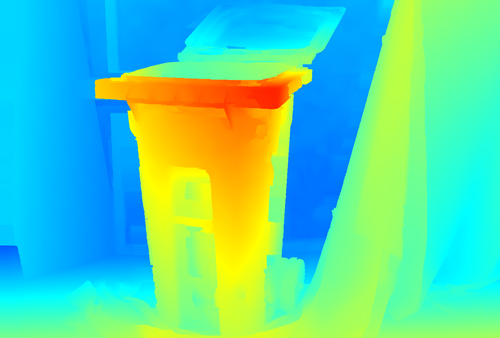} &
\includegraphics[width=0.258\columnwidth]{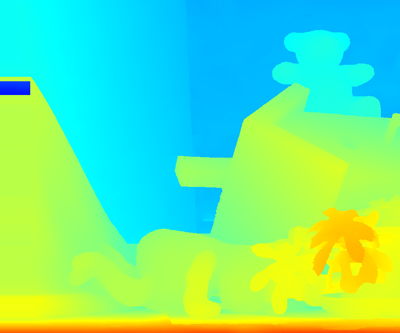} \\
MeshStr~\cite{zhang2015meshstereo}&
\includegraphics[width=0.31\columnwidth]{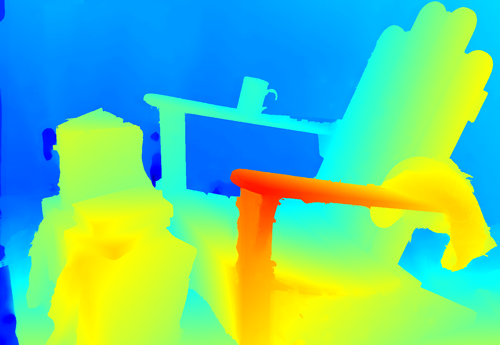} &
\includegraphics[width=0.32\columnwidth]{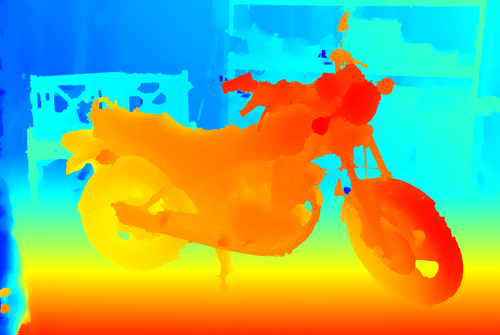} &
\includegraphics[width=0.313\columnwidth]{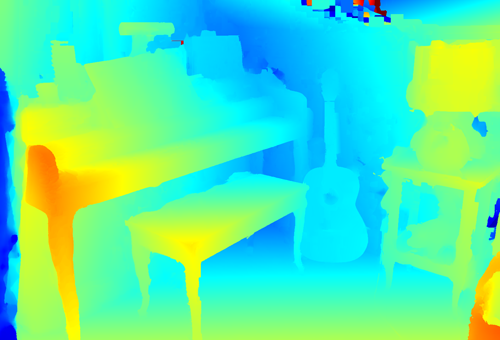} &
\includegraphics[width=0.318\columnwidth]{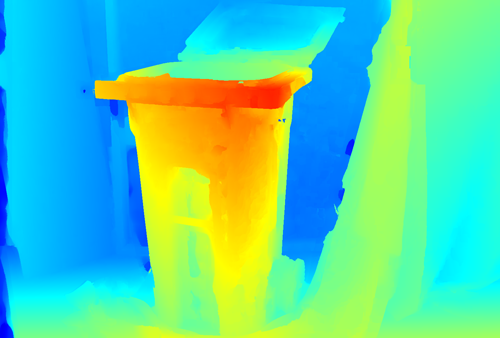} &
\includegraphics[width=0.258\columnwidth]{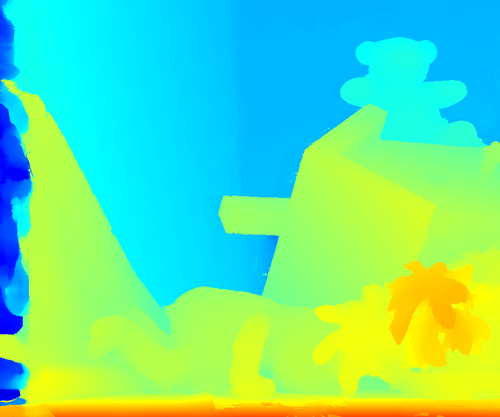} \\
NTDE~\cite{kim2016adaptive}&
\includegraphics[width=0.31\columnwidth]{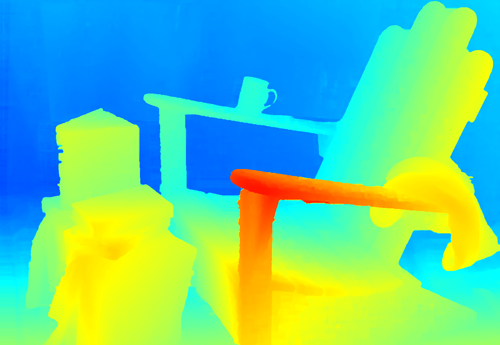} &
\includegraphics[width=0.32\columnwidth]{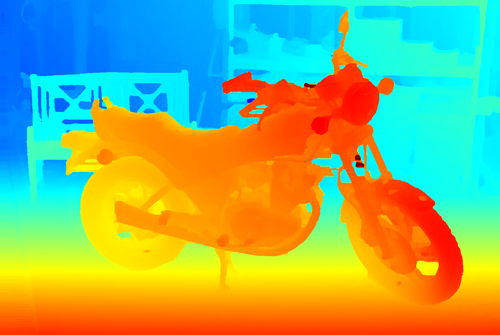} &
\includegraphics[width=0.313\columnwidth]{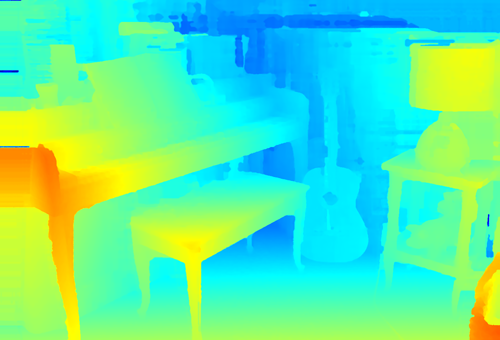} &
\includegraphics[width=0.318\columnwidth]{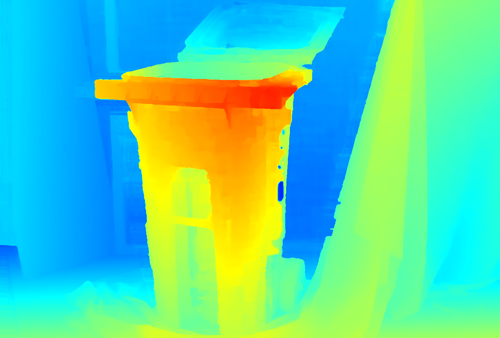} &
\includegraphics[width=0.258\columnwidth]{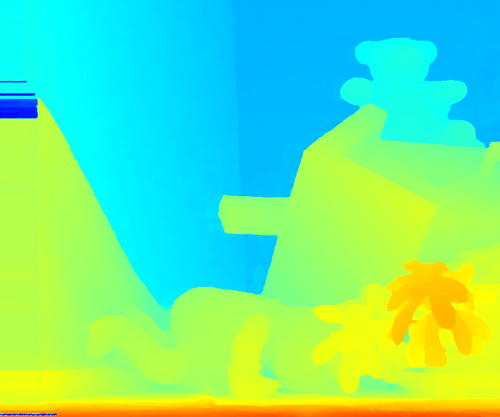} \\
& Adirondack & Motorcycle & PianoL & Recycle &Teddy\\
\end{tabular}
\caption{Resultant disparity maps of Adirondack, Motorcycle, PianoL,  Recycle, Teddy stereo images for visual  comparison. Here we compare our method with the other top five methods.
We take the example stereo images and the resultant disparity maps from the Middlebury benchmark V3 training set. }
\label{fg:disparity}
\end{center}
\end{figure*}
The experiments have been designed to verify the main contributions of the paper. They are divided into four parts where: 
\begin{itemize}
\item we compare the proposed approach to state-of-the-art;
\item we evaluate the advantage of the proposed sequential approach over non-sequential approach;
\item we analyze the advantage of using the OVOD model for energy  minimization;
\item we analyze the results of the proposed fully connected CRF model based on geodesic distance affinity implementation applied to the stereo matching problem.
\end{itemize}
In our experiments we use the Middlebury data set~\cite{scharstein2014high} (half-size). 

\subsection{Comparison to state-of-the-art methods}

Here we compare our method against the state-of-the-art methods on the Middlebury benchmark dataset. We consider the top five methods from the Middlebury benchmark table: LocalExp~\cite{taniai2016continuous}, 3DMST~\cite{li20173d}, PMSC~\cite{li2016pmsc}, MeshStereoExt~\cite{zhang2015meshstereo}, NTDE~\cite{kim2016adaptive}. Our algorithm achieves similar results as state-of-the-art for average (avrg), root-mean-squared (rms), A90, and A95 error metrics. Results are provided in Fig.~\ref{fg:table}. In this root-mean-squared error metric our method achieves second rank for the all pixels mask and first-second rank for the non-occluded pixels mask. Also results of 7 images with different masks for this metric achieve first rank in comparison with the second method LocalExp~\cite{taniai2016continuous} method that reaches only 3 first positions.

We also make Table~\ref{table:err1} to show the position of our algorithm among other top methods. Here the average errors over all the Middlebury data set images for all metrics are presented.  From this table we can see that for bad pixels metrics the results of our algorithm is not so good as for other error metrics, but still in the top ten methods for the all pixels mask. We think, that this is the effect of our scaling-up-down  procedure. Our main energy minimization process is severely restricted by the factor $M^3$ of the cost volume. To apply our method on the original image grid we have to increase memory usage and calculation time up to 125. Thus there is a trade-off between accuracy and computational resources. We also want to point out that the average best algorithm, LocalExp~\cite{taniai2016continuous}, needs to compute the additional right-to-left cost and the corresponding disparity map.  In contrast our OVOD approach uses only the left-to-right matching cost and we do not need to estimate the right-to-left disparity map.
On the other hand we obtain top position for the A95 and A99 error metrics. These metrics show that our method is very accurate for the majority of the pixels; the average error over the 95\% (and 99\%) of pixels with least error is very low.

To complete our comparison, we prepare Fig.~\ref{fg:table_time} for the run-time of the methods. One can see that our method is on the first position while all other methods are in inverse order in comparison with the results that are shown in Fig.~\ref{fg:table}, where the accuracy is considered. Thus, we can see that in general methods that are more accurate need more time to complete the same estimation task.  Considering the relative fastness of our method we can say that there are two reasons behind this fact: the OVOD approach uses only one cost volume, thereby reducing its run-time by almost a factor two. Furthermore, our method uses a cost-volume downscaling procedure that considerably decreases computational complexity. For example, if our algorithm works with the original image scale, the largest cost volume consists of 500 mega voxels. It means that we need approximately 50 GB RAM on  double floating-point precision. Here we count that every voxel corresponds to at least 8 message values. With our base scale factor that equals to 4 we can reduce the memory usage and the algorithm run-time by approximately 64 times. 

We prepare Fig.~\ref{fg:disparity} for visual comparison. In this figure we show the resultant disparity maps of Adirondack, Motorcycle, PianoL,  Recycle, Teddy obtained by the top stereo methods. 
 
\begin{table*}[tb]
\centering
\caption{Middlebury benchmark V3 for different error metrics. Here we compare our method with five other methods. Snapshot taken from website on December 15, 2017. Only average over all images of the full test data set is provided. The top part of this table with the all pixels mask and the bottom  the non-occluded mask. }
\label{table:err1}
\begin{tabular}{|c|c|c|c|c|c|c|c|c|c|c|}

	
\hline 
\rule{0cm}{0.35cm}
Error metrics: &  bad 0.5  &  bad 1 &  bad 2  &  bad 4 &  avrg   & rms &  A50  &  A90  &  A95 &  A99 \\[0.07cm]

 \hline 
 \hline 
 \rule{0cm}{0.35cm}	 
OVOD~(our)~~~~~~~~                             &$45.8_{~5}$ & $24.6_{~5}$  & $15.8_{~8}$ & $10.8_{~7}$   &  $5.19_{~2}$  &  $22.3_{~2}$ & $0.50_{~5}$ 
                                               &$5.60_{~5}$  & $\textbf{{27.1}}_{~1}$   & $\textbf{{108}}_{~1} $   \\ 
LocalExp~\cite{taniai2016continuous}~~~~~~     & $\textbf{{44.2}}_{~1}$ &  $\textbf{{21.0}}_{~1}$  &$\textbf{{11.7}}_{~1}$      &  $8.83_{~3}$  
                                               & $\textbf{{5.13}}_{~1}$ & $\textbf{{21.1}}_{~1} $ & $\textbf{{0.47}}_{~1} $ & $	5.27_{~4}$ 
                                               &$	31.6_{~6}$    & $109_{~2}$   \\ 
3DMST~\cite{li20173d}~~~~~~~~                         & $45.7_{~4}$ &$	22.0_{~2}$   & $	12.5_{~3}$  &  $8.81_{~2}$ &$	5.40_{~3}$    &$23.5_{~3}$  
                                                   &$0.49_{~4}$    & $	4.23_{~2}$ & $29.5_{~3}$   &  $114_{~3}$   \\ 
PMSC~\cite{li2016pmsc}~~~~~~~~~                       &$45.4_{~3}$& $	22.8_{~4}$  & $	13.6_{~4}$  &  $9.71_{~4}$ & $5.65_{~4}$   &  $	23.9_{~4}$ 
                                                   &  $0.49_{~3}$  &$5.20_{~3}$  &$31.4_{~5}$   & $115_{~4}$    \\ 
MeshStereoExt~\cite{zhang2015meshstereo}          &	$47.0_{~9}$	  &$24.8_{~6}$	   & $	15.7_{~7}$  &  $12.0_{~10}$ & $11.3_{~26}$   &  $	39.0_{~36}$ 
                                                  & $0.50_{~6}$   & $	26.4_{~20}$ & $	88.4_{~35}$   & $193_{~40}$    \\ 
NTDE~\cite{kim2016adaptive}~~~~~~~~~~~                  &$46.9_{~7}$&$	24.9_{~7}$  & $	15.3_{~6}$  &  $10.4_{~5}$ & $	6.20_{~7}$   &  $	26.1_{~7}$ 
                                               & $	0.51_{~7}$   & $	5.82_{~6}$ & $34.4_{~7}$   &  $	128_{~7}$   \\  [0.07cm]
\hline 
\hline 
 \rule{0cm}{0.35cm}	 
OVOD~(our)~~~~~~~~                             &$40.0_{~5}$ &$17.1_{~9}$  &$8.87_{~13}$	& $5.03_{~12}$ & $2.23_{~2}$  &  $\textbf{{12.9}}_{~1}$ 
                                               & $0.44_{~3}$   & $2.08_{~11}$ & $4.80_{~6}$  &  $\textbf{{52.6}}_{~1}$  \\ 
LocalExp~\cite{taniai2016continuous}~~~~~~     & $\textbf{{38.7}}_{~1}$ &  $\textbf{{13.9}}_{~1} $ & $\textbf{{5.43}}_{~1} $  &  $\textbf{{	3.69}}_{~1} $ 
                                               & $	2.24_{~3}$  &$13.4_{~4}$  &  $\textbf{{	0.43}}_{~1} $   & $1.55_{~3}$ & $4.81_{~7}$  &  $	55.3_{~3}$   \\ 
3DMST~\cite{li20173d}~~~~~~~~                         & $39.9_{~4}$ & $14.5_{~2}$  &$5.92_{~2}$    & $	3.72_{~2}$ & $\textbf{{2.17}}_{~1}$  & $13.3_{~3}$
                                                      &  $0.44_{~3}$   & $	1.53_{~2}$ &$\textbf{{3.47}}_{~1}$   & $55.0_{~2}$     \\ 
PMSC~\cite{li2016pmsc}~~~~~~~~~                       &$39.1_{~2}$& $14.8_{~3}$  &$	6.71_{~4}$    &$4.44_{~6}$  & $	2.26_{~4}$   &  $\textbf{{12.9}}_{~1}$ 
                                                      &  $\textbf{{	0.43}}_{~1} $   &$1.90_{~8}$  &  $5.27_{~8}$ &  $56.0_{~4}$    \\ 
MeshStereoExt~\cite{zhang2015meshstereo}          &	$	40.1_{~7}$	  & $	15.6_{~5}$  &$7.08_{~6}$    & $4.36_{~5}$ & $2.49_{~5}$   &$15.4_{~7}$
                                                     & $0.44_{~3}$    & $1.71_{~5}$ & $4.05_{~4}$  &   $67.3_{~7}$   \\ 
NTDE~\cite{kim2016adaptive}~~~~~~~~~~~                  &$40.4_{~9}$& $16.2_{~8}$  &$7.44_{~9}$    &$4.55_{~8}$  & $2.58_{~6}$   & $15.7_{~10}$
                                                      &  $0.44_{~3}$   & $	1.77_{~6}$ &  $4.74_{~5}$ &  $	65.4_{~6}$    \\  [0.07cm]
\hline
\end{tabular}
\end{table*}
\subsection{Comparison of sequential and non-sequential approaches}
\begin{figure*}[t]
\centering
\begin{tabular}{cc}
 \includegraphics[width=0.447\textwidth]{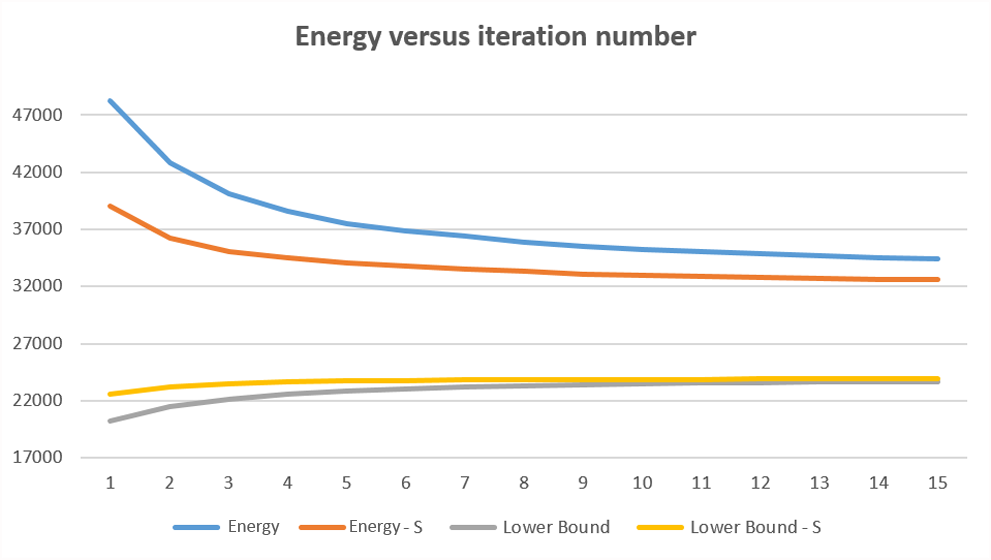} &
 \includegraphics[width=0.45\textwidth]{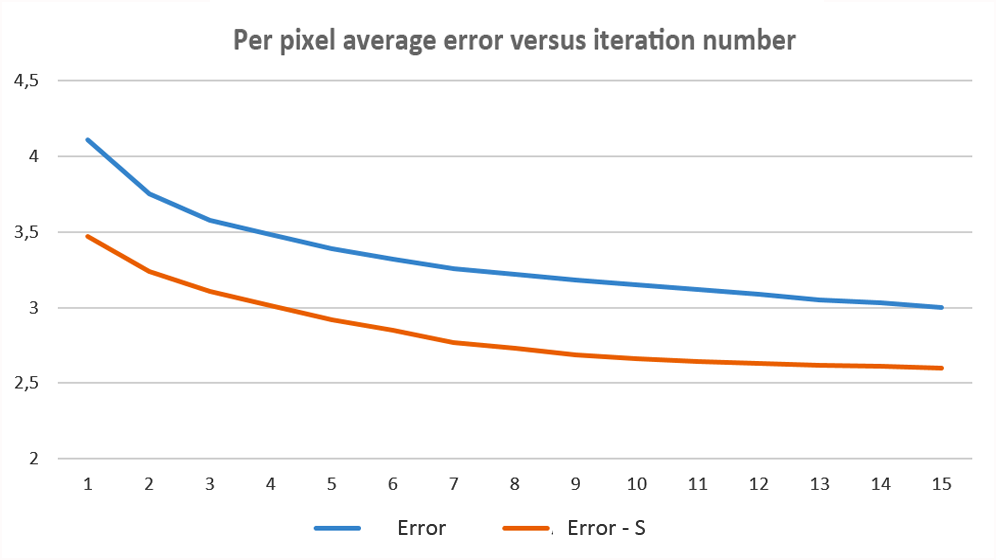}\\
(a)& (b)  \\
\end{tabular}
\caption {CRF energy convergence and corresponding accuracy for non-sequential and sequential realization of the proposed algorithm  for the Piano stereo image: 
(a) – Energy versus number of iteration; 
(b) – Per pixel average error versus number of iteration.}

\label{fg:EM_1}
\end{figure*}

For the next experiment on stereo matching we use the single Piano stereo image, because the average energy values over the full data set are not very representative due to different size, the number of disparity labels of an image and thus  different energy levels for each stereo image. We have verified that the convergence curves for the other images are almost identical to the Piano stereo image case. As cost we use the cost computed by~\cite{mozerov2015accurate}. In Fig.~\ref{fg:EM_1} the result of this experiment is provided. Here the energy value is calculated with Eq. ~(\ref{eq:BaseEM}) using intermediate disparity solutions on each iteration. Upper bound values are calculated by methodology proposed in~\cite{Kolmogorov06}.  Pixel accuracy improvement is presented in Fig.~\ref{fg:EM_1}(b). We can see that the sequential approach yields faster convergence and the gap between energy values and the lower bound estimation is less than for the non-sequential approach. 

\begin{figure}[t]
\centering
 \includegraphics[width=0.45\textwidth]{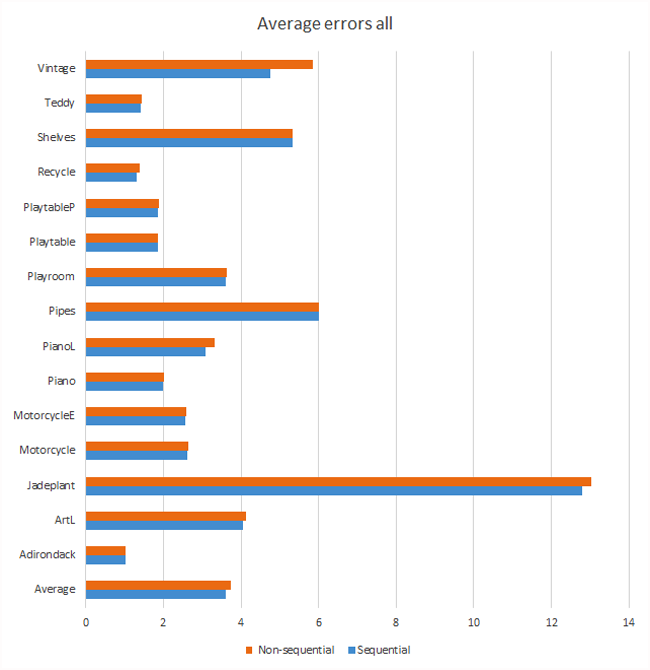} \\
\caption {The average errors (all pixels mask) obtained by the sequential (three iterations) and the non-sequential (six iterations) realization of the proposed algorithm. }
\label{fg:sq_1}
\end{figure}

Note that the convergence  of our energy minimization algorithm depends on the used cost. The CNN based cost~\cite{taniai2016continuous} which we use in our final experiments is more robust than the traditional cost (used in the previous experiment),  and in this case the minimization process converges considerably faster than, for example, with the cost taken from~\cite{mozerov2015accurate}. In our experiments with the Middlebury training data set for most stereo images results do not improve after three iterations. In this section we perform our second experiment  with the full training set. The sequential realization of our algorithm was applied with three iterations and non-sequential with six. The result of this experiment is given in Fig.~\ref{fg:sq_1}. We again can see that the sequential approach yields faster convergence and the accuracy of the sequential approach is higher even in the case, when the number of iterations is half as many as in the non-sequential case.   

\subsection{OVOD versus WTA}%
\begin{figure}[t]
\centering
 \includegraphics[width=0.45\textwidth]{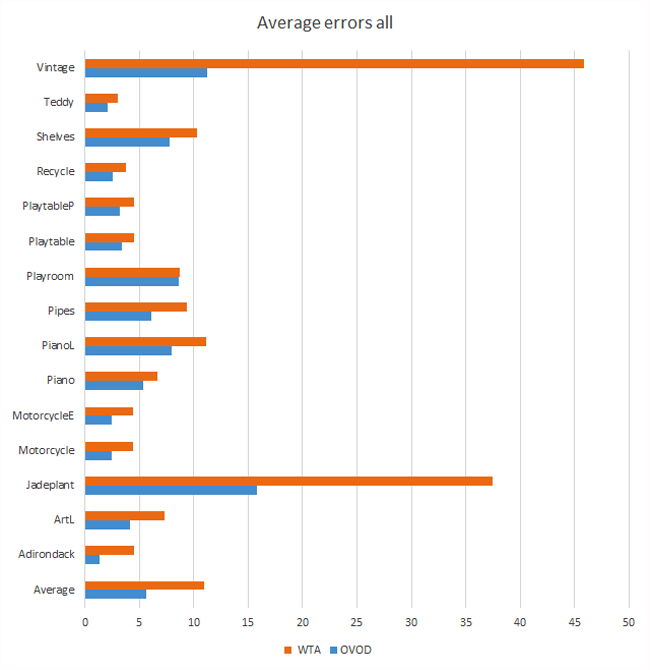} \\
\caption {The average errors (all pixels mask) obtained by our algorithm excluding any cost processing (the cost is considered as the BP marginals) with the standard WTA solution and with the proposed OVOD solution.}
\label{fg:wta_1}
\end{figure}
\begin{figure}[t]
\centering
 \includegraphics[width=0.45\textwidth]{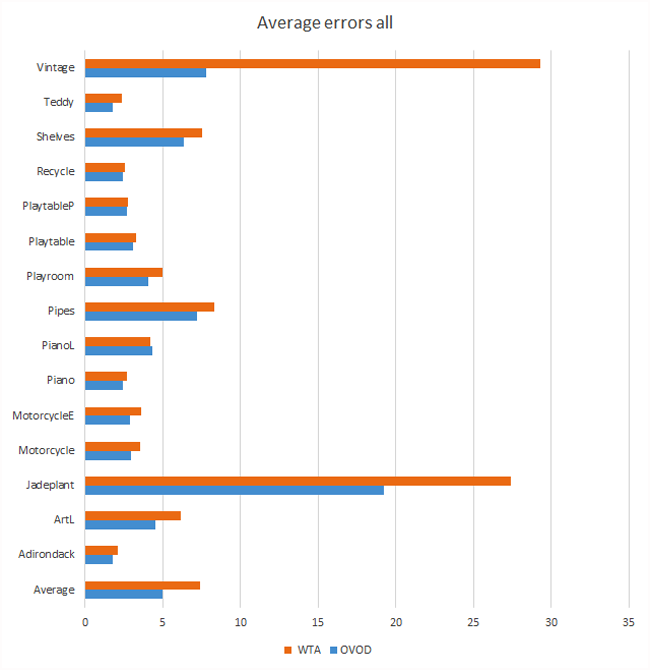} \\
\caption {The average errors (all pixels mask) obtained by our algorithm where the BP process is replaced by the cost filtering based on the geodesic distance filter (the approach similar to~\cite{Yang12}) with the standard WTA solution  and with the proposed OVOD solution.}
\label{fg:wta_2}
\end{figure}
\begin{figure}[t]
\centering
 \includegraphics[width=0.4\textwidth]{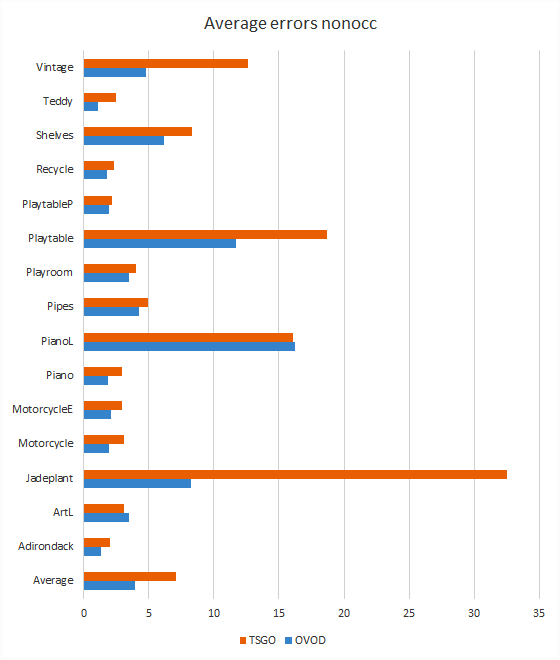}  \\

\caption {Comparison of our method with the TSGO algorithm~\cite{mozerov2015accurate} based on the same cost. The used metric is the average error in the non-occluded regions.}
\label{fg:crf_1}
\end{figure}
\begin{figure}[t]
\centering
 \includegraphics[width=0.4\textwidth]{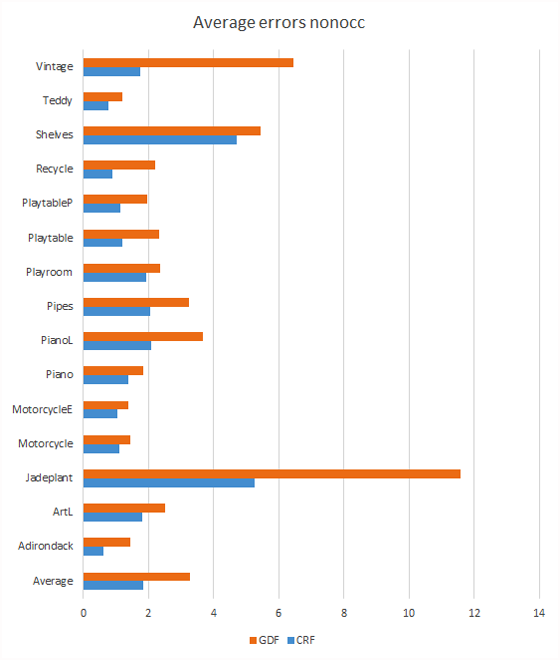}  \\

\caption {Comparison of our method with the cost filtering approach~\cite{Yang12} based on the CNN cost. The used metric is the average error in the non-occluded regions.}
\label{fg:crf_2}
\end{figure}

This experiment demonstrates the advantages of the proposed OVOD approach. We found that this approach can improve the matching results even for the case when the cost processing is not implemented; the cost is considered as the BP marginals. This case is illustrated in Fig.~\ref{fg:wta_1}. We can see that the OVOD solution is considerably more accurate than the solution obtained with the WTA operator, especially for metrics that count the occluded regions (all pixels). The drop of the error varies from 15\% on Teddy to 65\% on Vintage. Fig.~\ref{fg:wta_2}  demonstrates the advantages of the proposed OVOD approach in the case when the BP process in our method pipeline is replaced by the cost filtering based on the geodesic distance filter (an approach similar to~\cite{Yang12}). Again we can see that the proposed approach improves the accuracy of stereo matching. 

\subsection{Fully connected CRF models}
In this experiment we compare our method with two similar approach. The first approach, the TSGO algorithm~\cite{mozerov2015accurate}, uses the two step energy minimization technique and performs a left to right consistency check. To obtain a fair comparison we take the same cost as was used in~\cite{mozerov2015accurate}. Fig.~\ref{fg:crf_1} demonstrates that our approach leads to improved accuracy.  The average rank on the Middlebury of the TSGO is 47 for this data set, while the ranking result of the proposed method with the TSGO cost would be 22. 

The second approach we compare to is the cost filtering with geodesic distance affinity. This kind of filtering was used in~\cite{Yang12}.  To obtain a fair comparison we take the same CNN cost as was used in our main pipeline. Fig.~\ref{fg:crf_2} demonstrates that our approach leads to a significant drop of over 30\% on average. 

\section{Conclusions}\label{sec:conclusions}

Our work proposes two main contributions. First is a new theoretical contribution in the field of MAP: we derive a fully connected model for the BP approach.  We demonstrate that this model can be applied to the stereo matching problem and can improve the matching results in comparison with locally connected models. Our second contribution is the OVOD approach that allows to perform only one energy minimization process and avoids the cost calculation for the second view and the left-right check procedure. Application  of our algorithm on the Middlebury data set reach state-of-the-art results and for several error metrics even first rank position. We think that this results can be improved by applying the fine tune part of the 
local expansion moves algorithm~\cite{taniai2016continuous} as a post-processing step. 
\section*{Acknowledgements}
This work has been supported by the Spanish project TIN2015-65464-R, TIN2016-79717-R  (MINECO/FEDER) and the COST Action IC1307 iV\&L Net (European Network on Integrating Vision and Language), supported by COST (European Cooperation in Science and Technology). We acknowledge the CERCA Programme of Generalitat de Catalunya. We also acknowledge the generous GPU support from Nvidia.

{\small
\bibliographystyle{ieee}
\bibliography{egbib}
}

\end{document}